\documentclass[11pt,a4paper]{article}
\usepackage{authblk}
\usepackage{blindtext}
\usepackage[hyperref]{acl2018}
\usepackage{times}
\usepackage{latexsym}
\usepackage{qtree}
\usepackage{graphicx}
\usepackage{amsmath}
\usepackage{amsfonts,amssymb}
\usepackage{multirow}
\usepackage{multicol}
\usepackage{enumitem}
\usepackage{comment}
\usepackage{url}

\aclfinalcopy 
\def\aclpaperid{760} 

\newcommand{\matr}[1]{\mathbf{#1}}
\renewcommand{\vec}[1]{\mathbf{#1}}

\newcommand{\figref}[1]{Figure \ref{#1}}
\newcommand{\tabref}[1]{Table \ref{#1}}
\newcommand{\secref}[1]{Section \ref{#1}}

\usepackage{etoolbox}
\makeatletter
\patchcmd{\maketitle}
 {\def\@makefnmark}
 {\def\@makefnmark{}\def\useless@macro}
 {}{}
\makeatother

\title{Multi-Passage Machine Reading Comprehension \\ with Cross-Passage Answer Verification}

\author[1 *]{Yizhong Wang\thanks{\llap{\textsuperscript{*}}This work was done while the first author was doing internship at Baidu Inc.}}
\author[2]{Kai Liu}
\author[2]{Jing Liu}
\author[2]{Wei He}
\author[2]{\\Yajuan Lyu}
\author[2]{Hua Wu}
\author[1]{Sujian Li}
\author[2]{Haifeng Wang}
\affil[1]{Key Laboratory of Computational Linguistics, Peking University, MOE, China}
\affil[2]{Baidu Inc., Beijing, China}
\affil[ ]{\tt {\{yizhong, lisujian\}@pku.edu.cn, \{liukai20, liujing46, }}
\affil[ ]{\tt {hewei06, lvyajuan, wu\_hua, wanghaifeng\}@baidu.com}}

\date{}

\begin{document}
\maketitle

\begin{abstract}

Machine reading comprehension (MRC) on real web data usually requires the machine to answer a question by analyzing multiple passages retrieved by search engine. 
Compared with MRC on a single passage, multi-passage MRC is more challenging, since we are likely to get multiple confusing answer candidates from different passages.
To address this problem, we propose an end-to-end neural model that enables those answer candidates from different passages to verify each other based on their content representations.
Specifically, we jointly train three modules that can predict the final answer based on three factors: the answer boundary, the answer content and the cross-passage answer verification. 
The experimental results show that our method outperforms the baseline by a large margin and achieves the state-of-the-art performance on the English MS-MARCO dataset and the Chinese DuReader dataset, both of which are designed for MRC in real-world settings. 

\end{abstract}

\section{Introduction}
\label{introduction}


%


Machine reading comprehension (MRC), empowering computers with the ability to acquire knowledge and answer questions from textual data, is believed to be a crucial step in building a general intelligent agent \cite{cnn-dm-examination}. Recent years have seen rapid growth in the MRC community. With the release of various datasets, the MRC task has evolved from the early cloze-style test \cite{cnn-dm,cbt} to answer extraction from a single passage \cite{squad} and to the latest more complex question answering on web data \cite{marco, searchqa, dureader}. 

Great efforts have also been made to develop models for these MRC tasks 
, especially for the answer extraction on single passage \cite{match-lstm, bidaf, memen}. A significant milestone is that several MRC models have exceeded the performance of human annotators on the SQuAD dataset\footnote{https://rajpurkar.github.io/SQuAD-explorer/} \cite{squad}. However, this success on single Wikipedia passage is still not adequate, considering the ultimate goal of reading the whole web. Therefore, several latest datasets \cite{marco, dureader, searchqa} attempt to design the MRC tasks in more realistic settings by involving search engines. For each question, they use the search engine to retrieve multiple passages and the MRC models are required to read these passages in order to give the final answer.





One of the intrinsic challenges for such multi-passage MRC is that since all the passages are question-related but usually independently written, it's probable that multiple confusing answer candidates (correct or incorrect) exist. \tabref{tab:example} shows an example from MS-MARCO. We can see that all the answer candidates have semantic matching with the question while they are literally different and some of them are even incorrect. As is shown by \newcite{adversarial-examples}, these confusing answer candidates could be quite difficult for MRC models to distinguish. Therefore, special consideration is required for such multi-passage MRC problem.

\begin{table*}[htbp]
\small
\centering
\renewcommand{\arraystretch}{1.2}
\begin{tabular}{p{\textwidth}}
\hline
\textbf{Question: } What is the difference between a mixed and pure culture? \\
\hline
\textbf{Passages:} \\

[1] \textbf{A culture is a society's total way of living and a society is a group that live in a defined territory and participate in common culture.} While the answer given is in essence … true, societies originally form for the express purpose to enhance \ldots \\

[2] \ldots There has been resurgence in the economic system known as capitalism during the past two decades. 4. \textbf{The mixed economy is a balance between socialism and capitalism.} As a result, some institutions are owned and maintained by \ldots \\

[3] \textbf{A pure culture is one in which only one kind of microbial species is found whereas in mixed culture two or more microbial species formed colonies.} Culture on the other hand, is the lifestyle that the people in the country \ldots \\

[4] Best Answer: \textbf{A pure culture comprises a single species or strains. A mixed culture is taken from a source and may contain multiple strains or species.} A contaminated culture contains organisms that derived from some place \ldots \\

[5] \ldots It will be at that time when we can truly obtain a pure culture. \textbf{A pure culture is a culture consisting of only one strain.} You can obtain a pure culture by picking out a small portion of the mixed culture \ldots \\

[6] \textbf{A pure culture is one in which only one kind of microbial species is found whereas in mixed culture two or more microbial species formed colonies.} A pure culture is a culture consisting of only one strain. \ldots \\

$\cdots$ $\cdots$ \\
\hline
\textbf{Reference Answer:} A pure culture is one in which only one kind of microbial species is found whereas in mixed culture two or more microbial species formed colonies. \\
\hline
\end{tabular}
\caption{An example from MS-MARCO. The text in bold is the predicted answer candidate from each passage according to the boundary model. The candidate from [1] is chosen as the final answer by this model, while the correct answer is from [6] and can be verified by the answers from [3], [4], [5]. } \label{tab:example}
\end{table*}

In this paper, we propose to leverage the answer candidates from different passages to verify the final correct answer and rule out the noisy incorrect answers. 
Our hypothesis is that the correct answers could occur more frequently in those passages and usually share some commonalities, while incorrect answers are usually different from one another. The example in \tabref{tab:example} demonstrates this phenomenon. We can see that the answer candidates extracted from the last four passages are all valid answers to the question and they are semantically similar to each other, while the answer candidates from the other two passages are incorrect and there is no supportive information from other passages. As human beings usually compare the answer candidates from different sources to deduce the final answer, we hope that MRC model can also benefit from the cross-passage answer verification process.


The overall framework of our model is demonstrated in \figref{fig:architecture} , which consists of three modules. First, we follow the boundary-based MRC models \cite{bidaf,match-lstm} to find an answer candidate for each passage by identifying the start and end position of the answer (\figref{fig:content}). Second, we model the meanings of the answer candidates extracted from those passages and use the content scores to measure the quality of the candidates from a second perspective. Third, we conduct the answer verification by enabling each answer candidate to attend to the other candidates based on their representations. We hope that the answer candidates can collect supportive information from each other according to their semantic similarities and further decide whether each candidate is correct or not. Therefore, the final answer is determined by three factors: the boundary, the content and the cross-passage answer verification. The three steps are modeled using different modules, which can be jointly trained in our end-to-end framework. 


We conduct extensive experiments on the MS-MARCO \cite{marco} and DuReader \cite{dureader} datasets. The results show that our answer verification MRC model outperforms the baseline models by a large margin and achieves the state-of-the-art performance on both datasets. 



%
%
%

\section{Our Approach}
\label{approach}

\begin{figure*}[ht]
\centering
\includegraphics[width=\textwidth]{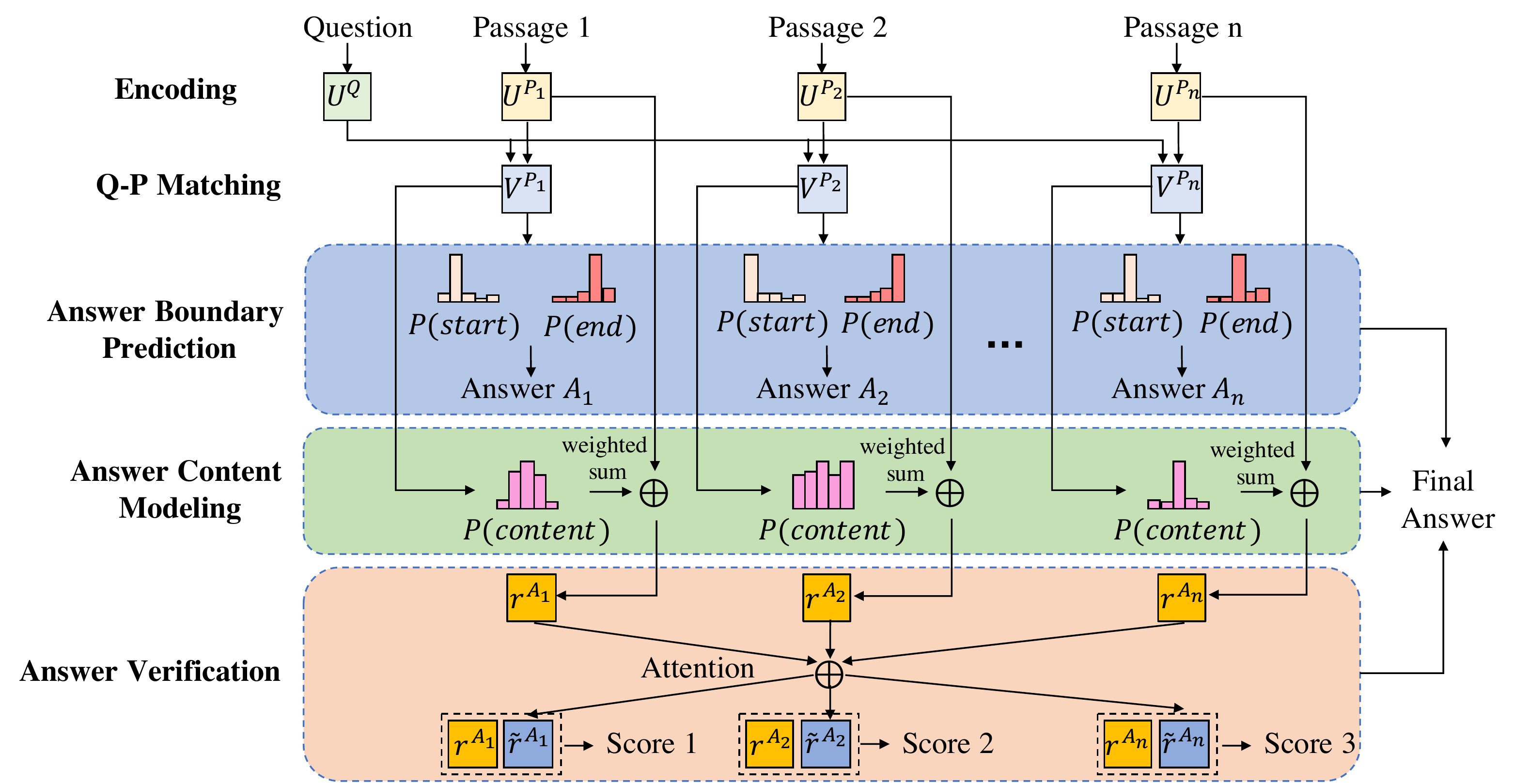}

\caption{Overview of our method for multi-passage machine reading comprehension}
\label{fig:architecture}
\end{figure*}

\figref{fig:architecture} gives an overview of our multi-passage MRC model
which is mainly composed of three modules including answer boundary prediction, answer content modeling and answer verification.
First of all, we need to model the question and passages. Following \newcite{bidaf}, we compute the question-aware representation for each passage (\secref{modeling}).
Based on this representation, we employ a Pointer Network \cite{pointer-net} to predict the start and end position of the answer in the module of answer boundary prediction (\secref{boundary}). 
At the same time, with the answer content model (\secref{content}), we estimate whether each word should be included in the answer and thus obtain the answer representations. 
Next, in the answer verification module (\secref{verification}), each answer candidate can attend to the other answer candidates to collect supportive information and we compute one score for each candidate to indicate whether it is correct or not according to the verification. 
The final answer is determined by not only the boundary but also the answer content and its verification score (\secref{train}).


\subsection{Question and Passage Modeling}
\label{modeling}

Given a question $\matr{Q}$ and a set of passages 
$\{\matr{P}_i\}$  retrieved by search engines, 
our task is to find the best concise answer to the question. First, we formally present the details of modeling the question and passages.

\paragraph{Encoding} We first map each word into the vector space by concatenating its word embedding and sum of its character embeddings. Then we employ bi-directional LSTMs (BiLSTM) to encode the question $\matr{Q}$ and passages $\{\matr{P}_i\}$ as follows:
\begin{align}
{\vec{u}}_t^Q & = \textrm{BiLSTM}_Q({\vec{u}}_{t-1}^Q, [{\vec{e}}_t^Q, {\vec{c}}_t^Q]) \\
{\vec{u}}_t^{P_i} & = \textrm{BiLSTM}_P({\vec{u}}_{t-1}^{P_i}, [{\vec{e}}_t^{P_i}, {\vec{c}}_t^{P_i}])
\end{align}
\noindent where ${\vec{e}}_t^Q$, ${\vec{c}}_t^Q$, ${\vec{e}}_t^{P_i}$, ${\vec{c}}_t^{P_i}$ are the word-level and character-level embeddings of the $t^{th}$ word. ${\vec{u}}_t^Q$ and ${\vec{u}}_t^{P_i}$ are the encoding vectors of the $t^{th}$ words in $\matr{Q}$ and $\matr{P}_i$ respectively.
Unlike previous work \cite{rnet} that simply concatenates all the passages, we process the passages independently at the encoding and matching steps. 

\paragraph{Q-P Matching} One essential step in MRC is to match the question with passages so that important information can be highlighted. We use the Attention Flow Layer \cite{bidaf} to conduct the Q-P matching in two directions. The similarity matrix $\matr{S} \in \mathbb{R}^{|\matr{Q}|
\times |\matr{P}_i|}$ between the question and passage $i$ is changed to a simpler version, where the similarity between the $t^{th}$ word in the question and the $k^{th}$ word in passage $i$ is computed as:
\begin{equation}
	\matr{S}_{t,k} = {{\vec{u}}_t^Q}^{\intercal} \cdot {\vec{u}}_k^{P_i}
\end{equation}

Then the context-to-question attention and question-to-context attention is applied strictly following \newcite{bidaf} to obtain the question-aware passage representation $\{{\vec{\tilde{u}}}_t^{P_i}\}$.
We do not give the details here due to space limitation. Next, another BiLSTM is applied in order to fuse the contextual information and get the new representation for each word in the passage, which is regarded as the match output:
\begin{equation}
	{\vec{v}}_t^{P_i} = \textrm{BiLSTM}_M({\vec{v}}_{t-1}^{P_i}, {\vec{\tilde{u}}}_t^{P_i})
\end{equation}

Based on the passage representations, we introduce the three main modules of our model.

\subsection{Answer Boundary Prediction}
\label{boundary}

To extract the answer span from passages, mainstream studies try to locate the boundary of the answer, which is called boundary model. Following \cite{match-lstm}, we employ Pointer Network \cite{pointer-net} to compute the probability of each word to be the start or end position of the span:
\begin{align}
	g_k^t &= {\vec{w}_{1}^{a}}^{\intercal} \tanh ( \matr{W}_{2}^a [\vec{v}_k^{P}, \vec{h}_{t-1}^{a}] ) \\
	{\alpha}_k^t &= \textrm{exp} (g_k^t) / \sum\nolimits_{j=1}^{|\matr{P}|} \textrm{exp} (g_j^t) \\
	\vec{c}_t &= \sum\nolimits_{k=1}^{|\matr{P}|} {\alpha}_k^t \vec{v}_k^{P} \\
	\vec{h}_t^a &= \textrm{LSTM} (\vec{h}_{t-1}^a, \vec{c}_t)
\end{align}

%

By utilizing the attention weights, the probability of the $k^{th}$ word in the passage to be the start and end position of the answer is obtained as ${\alpha}_k^1$ and ${\alpha}_k^2$. It should be noted that the pointer network is applied to the concatenation of all passages, which is denoted as $\textrm{P}$ so that the probabilities are comparable across passages. This boundary model can be trained by minimizing the negative log probabilities of the true start and end indices:
\begin{equation}
	\mathcal{L}_{boundary} = - \frac{1}{N} \sum_{i=1}^N (\log {\alpha}_{y_i^{1}}^1 + \log {\alpha}_{y_i^{2}}^2)
\end{equation}
\noindent where $N$ is the number of samples in the dataset and $y_i^{1}$, $y_i^{2}$ are the gold start and end positions.


\subsection{Answer Content Modeling}
\label{content}



Previous work employs the boundary model to find the text span with the maximum boundary score as the final answer. However, in our context, besides locating the answer candidates, we also need to model their meanings in order to conduct the verification. 
An intuitive method is to compute the representation of the answer candidates separately after extracting them, 
but it could be hard to train such model end-to-end.
Here, we propose a novel method that can obtain the representation of the answer candidates based on probabilities. 

Specifically, we change the output layer of the classic MRC model. Besides predicting the boundary probabilities for the words in the passages, we also predict whether each word should be included in the content of the answer. The content probability of the $k^{th}$ word is computed as:
\begin{align}
	p_k^c &= \textrm{sigmoid} ({\vec{w}_{1}^{c}}^{\intercal} \textrm{ReLU} (\matr{W}_{2}^c \vec{v}_k^{P_i}) )
\end{align}

Training this content model is also quite intuitive. We transform the boundary labels into a continuous segment, which means the words within the answer span will be labeled as 1 and other words will be labeled as 0. In this way, we define the loss function as the averaged cross entropy:
\begin{equation}
\begin{split}
	\mathcal{L}_{content} = & - \frac{1}{N} \frac{1}{|\textrm{P}|} \sum_{i=1}^N \sum_{j=1}^{|P|} [  y_k^c\log p_{k}^c \\ 
	& + (1-y_k^c)\log (1 - p_{k}^c)]
\end{split}
\end{equation}

The content probabilities provide another view to measure the quality of the answer in addition to the boundary. Moreover, with these probabilities, we can represent the answer from passage $i$ as a weighted sum of all the word embeddings in this passage:
\begin{align}
	\vec{r}^{A_i} = \frac{1}{|\matr{P}_{i}|}\sum\nolimits_{k=1}^{|\matr{P}_{i}|} p_k^c [{\vec{e}}_k^{P_i}, {\vec{c}}_k^{P_i}]
\end{align}

\subsection{Cross-Passage Answer Verification}
\label{verification}

The boundary model and the content model focus on extracting and modeling the answer within a single passage respectively, 
with little consideration of the cross-passage information. 
However, as is discussed in \secref{introduction}, there could be multiple answer candidates from different passages
and some of them may mislead the MRC model to make an incorrect prediction. It's necessary to aggregate the information from different passages and choose the best one from those candidates. Therefore, we propose a method to enable the answer candidates to exchange information and verify each other through the cross-passage answer verification process. 

Given the representation of the answer candidates from all passages $\{\vec{r}^{A_{i}}\}$, each answer candidate then attends to other candidates to collect supportive information via attention mechanism:
\begin{align}
	s_{i, j} &= 
	 \begin{cases}
 		0, & \text{if } i=j, \\
 		{\vec{r}^{A_i}}^{\intercal} \cdot \vec{r}^{A_j},  & \text{otherwise}	
 \end{cases}
 \\
	{\alpha}_{i, j} &= \textrm{exp} (s_{i, j}) / \sum\nolimits_{k=1}^{n} \textrm{exp} (s_{i, k}) \\
	\vec{\tilde{r}}^{A_i} &= \sum\nolimits_{j=1}^{n} {\alpha}_{i, j}\vec{r}^{A_j}
\end{align}

Here $\vec{\tilde{r}}^{A_{i}}$ is the collected verification information from other passages based on the attention weights. Then we pass it together with the original representation $\vec{r}^{A_{i}}$ to a fully connected layer:
\begin{align}
	g_{i}^v &= {\vec{w}^v}^{\intercal} [\vec{r}^{A_i}, \vec{\tilde{r}}^{A_i}, \vec{r}^{A_i} \odot \vec{\tilde{r}}^{A_i} ]
\end{align}
We further normalize these scores over all passages to get the verification score for answer candidate $A_i$:
\begin{equation}
	p_i^v = \textrm{exp} (g_i^v) / \sum\nolimits_{j=1}^{n} \textrm{exp} (g_j^v)
\end{equation}

In order to train this verification model, we take the answer from the gold passage as the gold answer. And the loss function can be formulated as the negative log probability of the correct answer:
\begin{equation}
	\mathcal{L}_{verify} = - \frac{1}{N} \sum_{i=1}^N \log p_{y_i^v}^{v}
\end{equation}
where $y_i^v$ is the index of the correct answer in all the answer candidates of the $i^{th}$ instance . 

\subsection{Joint Training and Prediction}
\label{train}

As is described above, we define three objectives for the reading comprehension model over multiple passages: 1. finding the boundary of the answer; 2. predicting whether each word should be included in the content; 3. selecting the best answer via cross-passage answer verification. According to our design, these three tasks can share the same embedding, encoding and matching layers. Therefore, we propose to train them together as multi-task learning \cite{multi-task}. The joint objective function is formulated as follows:
\begin{equation}
	\mathcal{L} = \mathcal{L}_{boundary} + \beta_{1} \mathcal{L}_{content} + \beta_{2} \mathcal{L}_{verify}
\end{equation}
where $\beta_1$ and $\beta_2$ are two hyper-parameters that control the weights of those tasks.

When predicting the final answer, we take the boundary score, content score and verification score into consideration. 
We first extract the answer candidate $A_i$ that has the maximum boundary score from each passage $i$. This boundary score is computed as the product of the start and end probability of the answer span. Then for each answer candidate $A_i$, we average the content probabilities of all its words as the content score of $A_i$. And we can also predict the verification score for $A_i$ using the verification model. Therefore, the final answer can be selected from all the answer candidates according to the product of these three scores.

\section{Experiments}
\label{experiments}

To verify the effectiveness of our model on multi-passage machine reading comprehension,
we conduct experiments on the MS-MARCO \cite{marco} and DuReader \cite{dureader} datasets. Our method achieves the state-of-the-art performance on both datasets.


\subsection{Datasets}
We choose the MS-MARCO and DuReader datasets to test our method, since both of them are designed from real-world search engines and involve a large number of passages retrieved from the web. One difference of these two datasets is that MS-MARCO mainly focuses on the English web data, while DuReader is designed for Chinese MRC. This diversity is expected to reflect the generality of our method. In terms of the data size, MS-MARCO contains 102023 questions, each of which is paired up with approximately 10 passages for reading comprehension. As for DuReader, it keeps the top-5 search results for each question and there are totally 201574 questions.  


\begin{table}[tbp]
\centering
\begin{tabular}{|c|c|c|}
\hline
                          & MS-MARCO & DuReader \\ \hline
Multiple Answers          &   9.93\%       &  67.28\%        \\
Multiple Spans &    40.00\%      &  56.38\%  \\\hline
\end{tabular}
\caption{Percentage of questions that have multiple valid answers or answer spans}
\label{tab:multi-answer}
\end{table}

One prerequisite for answer verification is that there should be multiple correct answers so that they can verify each other. Both the MS-MARCO and DuReader datasets require the human annotators to generate multiple answers if possible. \tabref{tab:multi-answer} shows the proportion of questions that have multiple answers. However, the same answer that occurs many times is treated as one single answer here. 
Therefore, we also report the proportion of questions that have multiple answer spans to match with the human-generated answers. A span is taken as valid if it can achieve F1 score larger than 0.7 compared with any reference answer. From these statistics, we can see that the phenomenon of multiple answers is quite common for both MS-MARCO and DuReader. These answers will provide strong signals for answer verification if we can leverage them properly.

\begin{table*}[t!]
\centering
\begin{tabular}{lcc}
\hline
Model           & ROUGE-L & BLEU-1 \\ \hline
FastQA\_Ext \cite{fastqa}      & 33.67   & 33.93  \\
Prediction \cite{match-lstm}  & 37.33   & 40.72  \\
ReasoNet   \cite{reasonet}     & 38.81   & 39.86  \\
R-Net      \cite{rnet}     & 42.89   & 42.22  \\
S-Net      \cite{snet}     & 45.23   & 43.78  \\ 
Our Model          & \textbf{46.15}        &  \textbf{44.47}      \\ \hline
S-Net      (Ensemble)     & 46.65   & 44.78  \\
Our Model (Ensemble) & \textbf{46.66}        &  \textbf{45.41}      \\ \hline
Human           & 47      & 46   \\ \hline
\end{tabular}
\caption{Performance of our method and competing models on the MS-MARCO test set}
\label{tab:marco-results}
\end{table*}

\subsection{Implementation Details}
\label{sec:implementation}

For MS-MARCO, we preprocess the corpus with the reversible tokenizer from Stanford CoreNLP \cite{stanford-corenlp} and we choose the span that achieves the highest ROUGE-L score with the reference answers as the gold span for training. We employ the 300-D pre-trained Glove embeddings \cite{glove} and keep it fixed during training. The character embeddings are randomly initialized with its dimension as 30. For DuReader, we follow the preprocessing described in \newcite{dureader}. 

We tune the hyper-parameters according to the validation performance on the MS-MARCO development set. The hidden size is set to be 150 and we apply $L2$ regularization with its weight as 0.0003. The task weights $\beta_{1}$, $\beta_{2}$ are both set to be 0.5. To train our model, we employ the Adam algorithm \cite{adam} with the initial learning rate as 0.0004 and the mini-batch size as 32. Exponential moving average is applied on all trainable variables with a decay rate 0.9999.

Two simple yet effective technologies are employed to improve the final performance on these two datasets respectively. For MS-MARCO, approximately 8\% questions have the answers as Yes or No, which usually cannot be solved by extractive approach \cite{snet}. We address this problem by training a simple Yes/No classifier for those questions with certain patterns (e.g., starting with ``is''). Concretely, we simply change the output layer of the basic boundary model so that it can predict whether the answer is ``Yes" or ``No". For DuReader, the retrieved document usually contains a large number of paragraphs that cannot be fed into MRC models directly \cite{dureader}. The original paper employs a simple a simple heuristic strategy to select a representative paragraph for each document, while we train a paragraph ranking model for this. We will demonstrate the effects of these two technologies later.

\subsection{Results on MS-MARCO}

\tabref{tab:marco-results} shows the results of our system and other state-of-the-art models on the MS-MARCO test set. We adopt the official evaluation metrics, including ROUGE-L \cite{rouge} and BLEU-1 \cite{bleu}. As we can see, for both metrics, our single model outperforms all the other competing models with an evident margin, which is extremely hard considering the near-human performance. If we ensemble the models trained with different random seeds and hyper-parameters, the results can be further improved and outperform the ensemble model in \newcite{snet}, especially in terms of the BLEU-1.

\subsection{Results on DuReader}

\begin{table}[tbp]
\centering
\begin{tabular}{|l|c|c|}
\hline
Model        & BLEU-4 & ROUGE-L \\ \hline
Match-LSTM   & 31.8   & 39.0  \\ 
BiDAF        & 31.9   & 39.2  \\
PR + BiDAF   & 37.55   & 41.81  \\ 
Our Model    & \textbf{40.97}   &  \textbf{44.18} \\ \hline
Human          & 56.1 & 57.4 \\ \hline
\end{tabular}
\caption{Performance on the DuReader test set}
\label{tab:dureader-results}
\end{table}

\begin{table}[tbp]
\centering
\begin{tabular}{|l|c|c|}
\hline
Model           & ROUGE-L & $\Delta$ \\ \hline
\textbf{Complete Model}      & \textbf{45.65}   &  \textbf{-} \\
 \- Answer Verification  &  44.38  & -1.27  \\
 \- Content Modeling   & 44.27   & -1.38  \\
 \- Joint Training  & 44.12   & -1.53         \\
 \- YesNo Classification &  41.87  &  -3.78 \\ 
\textbf{Boundary Baseline}   &  \textbf{38.95}  &  \textbf{-6.70} \\ \hline
\end{tabular}
\caption{Ablation study on MS-MARCO development set}
\label{tab:ablation}
\end{table}

The results of our model and several baseline systems on the test set of DuReader are shown in \tabref{tab:dureader-results}. The BiDAF and Match-LSTM models are provided as two baseline systems \cite{dureader}. Based on BiDAF, as is described in \secref{sec:implementation}, we tried a new paragraph selection strategy by employing a paragraph ranking (PR) model. We can see that this paragraph ranking can boost the BiDAF baseline significantly. Finally, we implement our system based on this new strategy, and our system (single model) achieves further improvement by a large margin.

\begin{table*}[tbp]
\centering
\small
\renewcommand{\arraystretch}{1.2}
\begin{tabular}{p{1.23\columnwidth}|ccc}
\hline
\textbf{Question:} What is the difference between a mixed and pure culture & \multicolumn{3}{c}{\bf{Scores}}\\ \hline
\multicolumn{1}{l|}{\bf{Answer Candidates:}} & \bf{Boundary} & \bf{Content} & \bf{Verification} \\
\textbf{[1]} A culture is a society's total way of living and a society is a group \ldots & \ensuremath{\bf{1.0\times10^{-2}}} & \ensuremath{\bf{1.0\times10^{-1}}} & \ensuremath{1.1\times10^{-1}} \\
\textbf{[2]} The mixed economy is a balance between socialism and capitalism. & \ensuremath{1.0\times10^{-4}} & \ensuremath{4.0\times10^{-2}} & \ensuremath{3.2\times10^{-2}} \\
\textbf{[3]} A pure culture is one in which only one kind of microbial species is \ldots & \ensuremath{5.5\times10^{-3}} & \ensuremath{7.7\times10^{-2}} & \ensuremath{1.2\times10^{-1}} \\
\textbf{[4]} A pure culture comprises a single species or strains. A mixed  \ldots & \ensuremath{2.7\times10^{-3}} & \ensuremath{8.1\times10^{-2}} & \ensuremath{1.3\times10^{-1}} \\
\textbf{[5]} A pure culture is a culture consisting of only one strain. & \ensuremath{5.8\times10^{-4}} & \ensuremath{7.9\times10^{-2}} & \ensuremath{5.1\times10^{-2}} \\
\textbf{[6]} \textbf{A pure culture is one in which only one kind of microbial species  \ldots} & \ensuremath{5.8\times10^{-3}} & \ensuremath{9.1\times10^{-2}} & \ensuremath{\bf{2.7\times10^{-1}}} \\
\multicolumn{1}{c|}{\bf{\ldots \ldots}} & \multicolumn{3}{c}{\bf{\ldots \ldots}}\\ \hline                               
\end{tabular}
\caption{Scores predicted by our model for the answer candidates shown in \tabref{tab:example}. Although the candidate [1] gets high boundary and content scores, the correct answer [6] is preferred by the verification model and is chosen as the final answer.}
\label{tab:case-study}
\end{table*}

\section{Analysis and Discussion}
\label{analysis}

\subsection{Ablation Study}


To get better insight into our system, we conduct in-depth ablation study on the development set of MS-MARCO, which is shown in \tabref{tab:ablation}. Following \newcite{snet}, we mainly focus on the ROUGE-L score that is averaged case by case. 

We first evaluate the answer verification by ablating the cross-passage verification model so that the verification loss and verification score will not be used during training and testing. Then we remove the content model in order to test the necessity of modeling the content of the answer. Since we don't have the content scores, we use the boundary probabilities instead to compute the answer representation for verification. Next, to show the benefits of joint training, we train the boundary model separately from the other two models. Finally, we remove the yes/no classification in order to show the real improvement of our end-to-end model compared with the baseline method that predicts the answer with only the boundary model. 

From \tabref{tab:ablation}, we can see that the answer verification makes a great contribution to the overall improvement, which confirms our hypothesis that cross-passage answer verification is useful for the multi-passage MRC. For the ablation of the content model, we analyze that it will not only affect the content score itself, but also violate the verification model since the content probabilities are necessary for the answer representation, which will be further analyzed in \secref{necessity}. Another discovery is that jointly training the three models can provide great benefits, which shows that the three tasks are actually closely related and can boost each other with shared representations at bottom layers. At last, comparing our method with the baseline, we achieve an improvement of nearly 3 points without the yes/no classification. This significant improvement proves the effectiveness of our approach.

\subsection{Case Study}

To demonstrate how each module of our model takes effect when predicting the final answer, we conduct a case study in \tabref{tab:case-study} with the same example that we discussed in \secref{introduction}. For each answer candidate, we list three scores predicted by the boundary model, content model and verification model respectively. 

On the one hand, we can see that these three scores generally have some relevance. For example, the second candidate is given lowest scores by all the three models. We analyze that this is because the models share the same encoding and matching layers at bottom level and this relevance guarantees that the content and verification models will not violate the boundary model too much. On the other hand, we also see that the verification score can really make a difference here when the boundary model makes an incorrect decision among the confusing answer candidates ([1], [3], [4], [6]). Besides, as we expected, the verification model tends to give higher scores for those answers that have semantic commonality with each other ([3], [4], [6]), which are all valid answers in this case. By multiplying the three scores, our model finally predicts the answer correctly.  

\subsection{Necessity of the Content Model}
\label{necessity}
\begin{figure*}[tb]
\centering
\includegraphics[width=\textwidth]{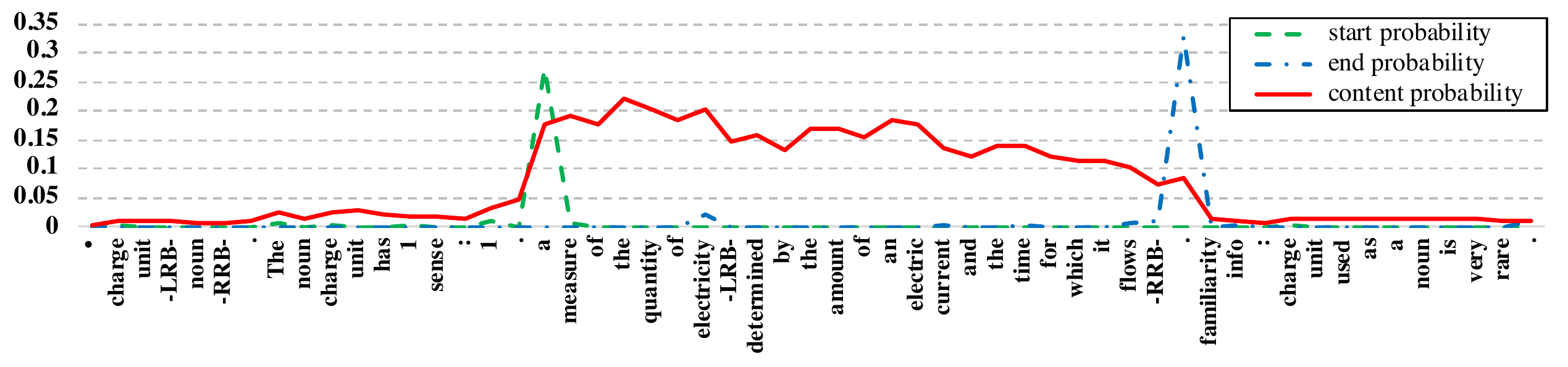}

\caption{The boundary probabilities and content probabilities for the words in a passage}
\label{fig:content}
\end{figure*}


In our model, we compute the answer representation based on the content probabilities predicted by a separate content model instead of directly using the boundary probabilities. 
We argue that this content model is necessary for our answer verification process.  \figref{fig:content} plots the predicted content probabilities as well as the boundary probabilities for a passage. We can see that the boundary and content probabilities capture different aspects of the answer. 
Since answer candidates usually have similar boundary words, if we compute the answer representation based on the boundary probabilities, it's difficult to model the real difference among different answer candidates.
On the contrary, with the content probabilities, we pay more attention to the content part of the answer, which can provide more distinguishable information for verifying the correct answer. Furthermore, the content probabilities can also adjust the weights of the words within the answer span so that unimportant words (e.g. ``and'' and ``.'') get lower weights in the final answer representation. We believe that this refined representation is also good for the answer verification process.

\section{Related Work}
\label{related}


Machine reading comprehension made rapid progress in recent years, especially for single-passage MRC task, such as SQuAD \cite{squad}. Mainstream studies \cite{bidaf, match-lstm, dcn} treat reading comprehension as extracting answer span from the given passage, which is usually achieved by predicting the start and end position of the answer.  We implement our boundary model similarly by employing the boundary-based pointer network \cite{match-lstm}. Another inspiring work is from \newcite{rnet}, where the authors  propose to match the passage against itself so that the representation can aggregate evidence from the whole passage. Our verification model adopts a similar idea. However, we collect information across passages and our attention is based on the answer representation, which is much more efficient than attention over all passages. For the model training, \newcite{dcn+} argues that  the boundary loss encourages exact answers at the cost of penalizing overlapping answers. Therefore they propose a mixed objective that incorporates rewards derived from word overlap. Our joint training approach has a similar function. By taking the content and verification loss into consideration, our model will give less loss for overlapping answers than those unmatched answers, and our loss function is totally differentiable.



Recently, we also see emerging interests in multi-passage MRC from both the academic \cite{searchqa, triviaqa} and industrial community \cite{marco, dureader}. Early studies \cite{reasonet, rnet} usually concat those passages and employ the same models designed for single-passage MRC. However, more and more latest studies start to design specific methods that can read multiple passages more effectively. In the aspect of passage selection, \newcite{r3} introduced a pipelined approach that rank the passages first and then read the selected passages for answering questions. \newcite{snet}  treats the passage ranking as an auxiliary task that can be trained jointly with the reading comprehension model. Actually, the target of our answer verification is very similar to that of the passage selection, while we pay more attention to the answer content and the answer verification process. Speaking of the answer verification, \newcite{evidence_aggregation} has a similar motivation to ours. They attempt to aggregate the evidence from different passages and choose the final answer from n-best candidates. However, they implement their idea as a separate reranking step after reading comprehension, while our answer verification is a component of the whole model that can be trained end-to-end.


\section{Conclusion}
\label{conclusion}

In this paper, we propose an end-to-end framework to tackle the multi-passage MRC task . We creatively design three different modules in our model, which can find the answer boundary, model the answer content and conduct cross-passage answer verification respectively. All these three modules can be trained with different forms of the answer labels and training them jointly can provide further improvement. The experimental results demonstrate that our model outperforms the baseline models by a large margin and achieves the state-of-the-art performance on two challenging datasets, both of which are designed for MRC on real web data.

\section*{Acknowledgments}
This work is supported by the National Basic Research Program of China (973 program, No. 2014CB340505) and Baidu-Peking University Joint Project.
We thank the Microsoft MSMARCO team for evaluating our results on the anonymous test set. We also thank Ying Chen, Xuan Liu and the anonymous reviewers for their constructive criticism of the manuscript.

\bibliography{ref}

\begin{thebibliography}{}
\expandafter\ifx\csname natexlab\endcsname\relax\def\natexlab#1{#1}\fi

\bibitem[{Chen et~al.(2016)Chen, Bolton, and Manning}]{cnn-dm-examination}
Danqi Chen, Jason Bolton, and Christopher~D. Manning. 2016.
\newblock A thorough examination of the cnn/daily mail reading comprehension
  task.
\newblock In {\em Proceedings of the 54th Annual Meeting of the Association for
  Computational Linguistics, {ACL} 2016, August 7-12, 2016, Berlin, Germany,
  Volume 1: Long Papers\/}.

\bibitem[{Dunn et~al.(2017)Dunn, Sagun, Higgins, G{\"{u}}ney, Cirik, and
  Cho}]{searchqa}
Matthew Dunn, Levent Sagun, Mike Higgins, V.~Ugur G{\"{u}}ney, Volkan Cirik,
  and Kyunghyun Cho. 2017.
\newblock Searchqa: {A} new q{\&}a dataset augmented with context from a search
  engine.
\newblock {\em arXiv preprint arXiv:1704.05179\/} .

\bibitem[{He et~al.(2017)He, Liu, Lyu, Zhao, Xiao, Liu, Wang, Wu, She, Liu, Wu,
  and Wang}]{dureader}
Wei He, Kai Liu, Yajuan Lyu, Shiqi Zhao, Xinyan Xiao, Yuan Liu, Yizhong Wang,
  Hua Wu, Qiaoqiao She, Xuan Liu, Tian Wu, and Haifeng Wang. 2017.
\newblock Dureader: a chinese machine reading comprehension dataset from
  real-world applications.
\newblock {\em arXiv preprint arXiv:1711.05073\/} .

\bibitem[{Hermann et~al.(2015)Hermann, Kocisk{\'{y}}, Grefenstette, Espeholt,
  Kay, Suleyman, and Blunsom}]{cnn-dm}
Karl~Moritz Hermann, Tom{\'{a}}s Kocisk{\'{y}}, Edward Grefenstette, Lasse
  Espeholt, Will Kay, Mustafa Suleyman, and Phil Blunsom. 2015.
\newblock Teaching machines to read and comprehend.
\newblock In {\em Advances in Neural Information Processing Systems 28: Annual
  Conference on Neural Information Processing Systems 2015\/}.

\bibitem[{Hill et~al.(2015)Hill, Bordes, Chopra, and Weston}]{cbt}
Felix Hill, Antoine Bordes, Sumit Chopra, and Jason Weston. 2015.
\newblock The goldilocks principle: Reading children's books with explicit
  memory representations.
\newblock {\em arXiv preprint arXiv:1511.02301\/} .

\bibitem[{Jia and Liang(2017)}]{adversarial-examples}
Robin Jia and Percy Liang. 2017.
\newblock Adversarial examples for evaluating reading comprehension systems.
\newblock In {\em Proceedings of the 2017 Conference on Empirical Methods in
  Natural Language Processing, {EMNLP} 2017, Copenhagen, Denmark, September
  9-11, 2017\/}. pages 2021--2031.

\bibitem[{Joshi et~al.(2017)Joshi, Choi, Weld, and Zettlemoyer}]{triviaqa}
Mandar Joshi, Eunsol Choi, Daniel Weld, and Luke Zettlemoyer. 2017.
\newblock Triviaqa: A large scale distantly supervised challenge dataset for
  reading comprehension.
\newblock In {\em Proceedings of the 55th Annual Meeting of the Association for
  Computational Linguistics\/}. volume~1, pages 1601--1611.

\bibitem[{Kingma and Ba(2014)}]{adam}
Diederik~P. Kingma and Jimmy Ba. 2014.
\newblock Adam: {A} method for stochastic optimization.
\newblock {\em arXiv preprint arXiv:1412.6980\/} .

\bibitem[{Lin(2004)}]{rouge}
Chin-Yew Lin. 2004.
\newblock Rouge: A package for automatic evaluation of summaries.
\newblock {\em Text Summarization Branches Out\/} .

\bibitem[{Manning et~al.(2014)Manning, Surdeanu, Bauer, Finkel, Bethard, and
  McClosky}]{stanford-corenlp}
Christopher~D. Manning, Mihai Surdeanu, John Bauer, Jenny Finkel, Steven~J.
  Bethard, and David McClosky. 2014.
\newblock The stanford corenlp natural language processing toolkit.
\newblock In {\em Association for Computational Linguistics (ACL) System
  Demonstrations\/}. pages 55--60.

\bibitem[{Nguyen et~al.(2016)Nguyen, Rosenberg, Song, Gao, Tiwary, Majumder,
  and Deng}]{marco}
Tri Nguyen, Mir Rosenberg, Xia Song, Jianfeng Gao, Saurabh Tiwary, Rangan
  Majumder, and Li~Deng. 2016.
\newblock {MS} {MARCO:} {A} human generated machine reading comprehension
  dataset.
\newblock In {\em Proceedings of the Workshop on Cognitive Computation:
  Integrating neural and symbolic approaches 2016 co-located with the 30th
  Annual Conference on Neural Information Processing Systems {(NIPS} 2016)\/}.

\bibitem[{Pan et~al.(2017)Pan, Li, Zhao, Cao, Cai, and He}]{memen}
Boyuan Pan, Hao Li, Zhou Zhao, Bin Cao, Deng Cai, and Xiaofei He. 2017.
\newblock Memen: Multi-layer embedding with memory networks for machine
  comprehension.
\newblock {\em arXiv preprint arXiv:1707.09098\/} .

\bibitem[{Papineni et~al.(2002)Papineni, Roukos, Ward, and Zhu}]{bleu}
Kishore Papineni, Salim Roukos, Todd Ward, and Wei{-}Jing Zhu. 2002.
\newblock Bleu: a method for automatic evaluation of machine translation.
\newblock In {\em Proceedings of the 40th Annual Meeting of the Association for
  Computational Linguistics, July 6-12, 2002, Philadelphia, PA, {USA.}\/}.
  pages 311--318.

\bibitem[{Pennington et~al.(2014)Pennington, Socher, and Manning}]{glove}
Jeffrey Pennington, Richard Socher, and Christopher~D. Manning. 2014.
\newblock Glove: Global vectors for word representation.
\newblock In {\em Empirical Methods in Natural Language Processing (EMNLP)\/}.
  pages 1532--1543.

\bibitem[{Rajpurkar et~al.(2016)Rajpurkar, Zhang, Lopyrev, and Liang}]{squad}
Pranav Rajpurkar, Jian Zhang, Konstantin Lopyrev, and Percy Liang. 2016.
\newblock Squad: 100, 000+ questions for machine comprehension of text.
\newblock In {\em Proceedings of the 2016 Conference on Empirical Methods in
  Natural Language Processing, {EMNLP} 2016\/}.

\bibitem[{Ruder(2017)}]{multi-task}
Sebastian Ruder. 2017.
\newblock An overview of multi-task learning in deep neural networks.
\newblock {\em arXiv preprint arXiv:1706.05098\/} .

\bibitem[{Seo et~al.(2016)Seo, Kembhavi, Farhadi, and Hajishirzi}]{bidaf}
Min~Joon Seo, Aniruddha Kembhavi, Ali Farhadi, and Hannaneh Hajishirzi. 2016.
\newblock Bidirectional attention flow for machine comprehension.
\newblock {\em arXiv preprint arXiv:1611.01603\/} .

\bibitem[{Shen et~al.(2017)Shen, Huang, Gao, and Chen}]{reasonet}
Yelong Shen, Po{-}Sen Huang, Jianfeng Gao, and Weizhu Chen. 2017.
\newblock Reasonet: Learning to stop reading in machine comprehension.
\newblock In {\em Proceedings of the 23rd {ACM} {SIGKDD} International
  Conference on Knowledge Discovery and Data Mining, Halifax, NS, Canada,
  August 13 - 17, 2017\/}. pages 1047--1055.

\bibitem[{Tan et~al.(2017)Tan, Wei, Yang, Lv, and Zhou}]{snet}
Chuanqi Tan, Furu Wei, Nan Yang, Weifeng Lv, and Ming Zhou. 2017.
\newblock S-net: From answer extraction to answer generation for machine
  reading comprehension.
\newblock {\em arXiv preprint arXiv:1706.04815\/} .

\bibitem[{Vinyals et~al.(2015)Vinyals, Fortunato, and Jaitly}]{pointer-net}
Oriol Vinyals, Meire Fortunato, and Navdeep Jaitly. 2015.
\newblock Pointer networks.
\newblock In {\em Advances in Neural Information Processing Systems 28: Annual
  Conference on Neural Information Processing Systems 2015, December 7-12,
  2015, Montreal, Quebec, Canada\/}. pages 2692--2700.

\bibitem[{Wang and Jiang(2016)}]{match-lstm}
Shuohang Wang and Jing Jiang. 2016.
\newblock Machine comprehension using match-lstm and answer pointer.
\newblock {\em arXiv preprint arXiv:1608.07905\/} .

\bibitem[{Wang et~al.(2017{\natexlab{a}})Wang, Yu, Guo, Wang, Klinger, Zhang,
  Chang, Tesauro, Zhou, and Jiang}]{r3}
Shuohang Wang, Mo~Yu, Xiaoxiao Guo, Zhiguo Wang, Tim Klinger, Wei Zhang, Shiyu
  Chang, Gerald Tesauro, Bowen Zhou, and Jing Jiang. 2017{\natexlab{a}}.
\newblock R{\textdollar}{\^{}}3{\textdollar}: Reinforced reader-ranker for
  open-domain question answering.
\newblock {\em arXiv preprint arXiv:1709.00023\/} .

\bibitem[{Wang et~al.(2017{\natexlab{b}})Wang, Yu, Jiang, Zhang, Guo, Chang,
  Wang, Klinger, Tesauro, and Campbell}]{evidence_aggregation}
Shuohang Wang, Mo~Yu, Jing Jiang, Wei Zhang, Xiaoxiao Guo, Shiyu Chang, Zhiguo
  Wang, Tim Klinger, Gerald Tesauro, and Murray Campbell. 2017{\natexlab{b}}.
\newblock Evidence aggregation for answer re-ranking in open-domain question
  answering.
\newblock {\em arXiv preprint arXiv:1711.05116\/} .

\bibitem[{Wang et~al.(2017{\natexlab{c}})Wang, Yang, Wei, Chang, and
  Zhou}]{rnet}
Wenhui Wang, Nan Yang, Furu Wei, Baobao Chang, and Ming Zhou.
  2017{\natexlab{c}}.
\newblock Gated self-matching networks for reading comprehension and question
  answering.
\newblock In {\em Proceedings of the 55th Annual Meeting of the Association for
  Computational Linguistics, {ACL} 2017, Vancouver, Canada, July 30 - August 4,
  Volume 1: Long Papers\/}.

\bibitem[{Weissenborn et~al.(2017)Weissenborn, Wiese, and Seiffe}]{fastqa}
Dirk Weissenborn, Georg Wiese, and Laura Seiffe. 2017.
\newblock Making neural {QA} as simple as possible but not simpler.
\newblock In {\em Proceedings of the 21st Conference on Computational Natural
  Language Learning (CoNLL 2017), Vancouver, Canada, August 3-4, 2017\/}. pages
  271--280.

\bibitem[{Xiong et~al.(2016)Xiong, Zhong, and Socher}]{dcn}
Caiming Xiong, Victor Zhong, and Richard Socher. 2016.
\newblock Dynamic coattention networks for question answering.
\newblock {\em arXiv preprint arXiv:1611.01604\/} .

\bibitem[{Xiong et~al.(2017)Xiong, Zhong, and Socher}]{dcn+}
Caiming Xiong, Victor Zhong, and Richard Socher. 2017.
\newblock {DCN+:} mixed objective and deep residual coattention for question
  answering.
\newblock {\em arXiv preprint arXiv:1711.00106\/} .

\end{thebibliography}
\bibliographystyle{acl_natbib}

\end{document}